\documentclass[a4paper,twoside]{article}

\usepackage{epsfig}
\usepackage{amsmath}
\usepackage[linesnumbered]{algorithm2e}

\usepackage{gnuplot-lua-tikz}
\usepackage{gnuplottex}
\usepackage{subcaption}
\usepackage{url}
\usepackage{makecell }
\usepackage{microtype}
\usepackage{booktabs}
\usepackage{tikz}
\usetikzlibrary{shapes, arrows.meta, positioning, calc}
\usepackage[capitalise, noabbrev]{cleveref}
\usepackage[style=apa,autocite=footnote,maxcitenames=1,minnames=1,maxalphanames=4,minalphanames=3,natbib=true,url=false, doi=false]{biblatex}
\usepackage{ntheorem}
\theoremseparator{:}
\newtheorem{hyp}{Hypothesis}

\usepackage{SCITEPRESS}     

\DefineBibliographyStrings{english}{%
    andothers = {et\addabbrvspace al\adddot}
}
\newcommand{\projectName}{TEAM}

\newcommand{\compagny}{Schindler}

\begin{document}

\title{\projectName{}: a parameter-free algorithm to teach collaborative robots motions from user demonstrations.}

\author{\authorname{Lorenzo Panchetti\sup{1}\orcidAuthor{0009-0004-9657-7249}, Jianhao Zheng\sup{1}\orcidAuthor{0000-0003-4430-3049}, Mohamed Bouri\sup{1}\orcidAuthor{0000-0003-1083-3180}, and Malcolm Mielle\sup{2}\orcidAuthor{0000-0002-3079-0512}}
    \affiliation{\sup{1}École polytechnique fédérale de Lausanne (EPFL), Lausanne, Switzerland}
    \affiliation{\sup{2}Schindler AG, EPFL Lab, Lausanne, Switzerland}
    \email{\{lorenzo.panchetti, malcolm.mielle\}@schindler.com, jianhaozheng1@gmail.com, mohamed.bouri@epfl.com}
}
%

\abstract{
    Learning from demonstrations (LfD) enables humans to easily teach collaborative robots (cobots) new motions that can be generalized to new task configurations without retraining.
    However, state-of-the-art LfD methods require manually tuning intrinsic parameters and have rarely been used in industrial contexts without experts.
    We propose a parameter-free LfD method based on probabilistic movement primitives, where parameters are determined using Jensen-Shannon divergence and Bayesian optimization, and users do not have to perform manual parameter tuning.
    The cobot's precision in reproducing learned motions, and its ease of teaching and use by non-expert users are evaluated in two field tests.
    In the first field test, the cobot works on elevator door maintenance.
    In the second test, three factory workers teach the cobot tasks useful for their daily workflow.
    Errors between the cobot and target joint angles are insignificant---at worst $0.28$ deg---and the motion is accurately reproduced---GMCC score of 1.
    Questionnaires completed by the workers highlighted the method's ease of use and the accuracy of the reproduced motion.
    Public implementation of our method and datasets are made available online.
}

\onecolumn \maketitle \normalsize \setcounter{footnote}{0} \vfill

\section{\uppercase{Introduction}}

Collaborative robots (cobots) are built to improve society by helping people without replacing them.
To become an integrated part of our work, human workers must be able to teach cobots new tasks in a short time, making the robot a new tool in their toolbox.
However,
programming the cobot is most of the time done by experts and cobots cannot adapt to new task configurations, instead repeating learned patterns.

Learning from demonstration (LfD) (\cite{Rana2020BenchmarkFS})---a branch of learning focused on skill transfer and generalization through a set of demonstrations---enables cobots to learn and adapt motions from a set of demonstrations.
State-of-the-art LfD methods require either manually tuning intrinsic parameters or a large amount of data, and have thus rarely been used in industrial contexts without experts, since manual tuning and data collection are time-consuming and error-prone.
In this paper, we present \projectName{} (\underline{te}ach a robot \underline{a}rm to \underline{m}ove), a novel method to learn from demonstrations \emph{without manual tuning of intrinsic parameters} during training.

The main contributions of this paper are:

\begin{itemize}
    \item A parameter-free framework to learn motions from a set of demonstrations, using a generative model to find a generalized trajectory, and attractor landscapes to reproduce the motion between different start and target joint angles.
    \item An optimization strategy of the attractor landscape's intrinsic parameters through Bayesian optimization.
    \item Improvement on the selection of the number of Gaussian Mixture Models through a series of one-tailed Welch's t-tests, based on the Jensen-Shannon divergence.
    \item Experimental validation of \projectName{} in two field tests showing that our method can be used by non-expert robot users.
\end{itemize}

\begin{figure}[t]
    \centering
    \scalebox{0.75}{\input{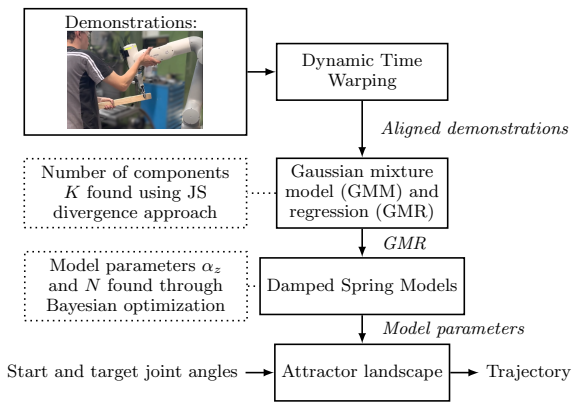}}
    \caption{
        A set of demonstrations is recorded by a user and the motion is generalized through GMR.
        The system is modelled as a set of damped spring models that generalize the motion.
        Given start and target joint angles, model's parameters are used to generate new trajectories reproducing the motion taught by demonstrations.
        All system parameters are automatically optimized, and no expert knowledge is needed.
    }
    \label{sec:introduction:flowchart}
\end{figure}

A complete overview of the methodology is shown in \cref{sec:introduction:flowchart}.
\section{\uppercase{Related work}}


\textcite{Rana2020BenchmarkFS} present a large-scale study benchmarking the performance of motion-based LfD approaches
and show that Probabilistic Movement Primitives (ProMP) (\cite{Paraschos2013ProbabilisticMP}) methods are the most consistent on tasks with positional constraints.
ProMP is a general probabilistic framework for learning movement primitives that allows new operations, including conditioning and adaptation to changed task variables.

\textcite{Calinon2007OnLR} fit a mixture of Gaussians on a set of demonstrations and generalize the motion through Gaussian Mixture Regression (GMR)(\cite{Cohn1996ActiveLW}).
Trajectories are computed by optimizing an imitation performance metric.
However, joint configurations are not constrained to the demonstration space, which can lead to the exploration of unsafe areas.

\textcite{Kulak2021ActiveLO} propose to use Bayesian Gaussian mixture models to learn ProMP.
While their method reduces the number of demonstrations needed to learn a representation with generalization capabilities, the method parameters must be manually set for all experiments.

\textcite{Schaal2006, Ijspeert2013Nonlinear} use Dynamical Movement Primitives (DMP) to model complex motions through nonlinear dynamical systems.
DMP is scale and temporal invariant, convergence is proven, but the parameters of the system must be manually tuned.

\textcite{Pervez2018LearningTD} propose a method that generalizes motion outside the demonstrated task space.
Each demonstration is associated with a dynamical system, and learning is formulated as a density estimation problem.
However, parameters must be set empirically for all dynamical systems.


Recent works have leveraged advances in deep learning.
To tackle the challenging problem of model collapse, \textcite{Zhou2020MovementPL} propose using a mixture density network (MDN) that takes task parameters as input and provides a Gaussian mixture model (GMM) of the MP parameters.
During training, their work introduces an entropy cost to achieve a more balanced association of demonstrations to GMM components.

\textcite{DeepMPRidge} propose to train a neural network to output the parameters of the DMP model from an image, before learning the associated forcing term.
\textcite{deep-MPPervez} use deep neural networks to learn the forcing terms of the DMP model for vision-based robot control.
Both methods involve a convolutional neural network learning task-specific features from camera images.
%
\textcite{breast} estimate the correlation between visual information and ProMP weights for reach-to-palpate motion.
The average error in task space is around 3 to 5 centimeters which is too high for our application.
\textcite{Quantao2022MPR} use reinforcement learning to learn a latent action space representing the skill embedding from demonstrated trajectories for each prior task.
\textcite{Tosatto2020ContextualLO} provide a complete framework for sample-efficient off-policy RL optimization of MP for robot learning of high-dimensional manipulation skills.
All methods based on deep or reinforcement learning require a large amount of data.
E.g., \textcite{breast} show the robot the reach-to-palpate motion 500 times, \textcite{deep-MPPervez} acquire 50 demonstrations for a single task, and \textcite{Quantao2022MPR} uses around 80K trajectories.

\section{\uppercase{Method}}

\subsection{Overview}

To learn a motion, a set of demonstrations is first collected by the user.
In our work, the cobot is taught by manual guidance---see the image in the demonstration box of \cref{sec:introduction:flowchart}.
A demonstration stores the cobot's joint angles recorded while the cobot is shown the task and the robot is controlled in joint space to avoid singularities during the motion of the redundant robot arm.
As in previous work by \textcite{Calinon2007OnLR}, demonstrations are aligned in time using dynamic time warping (\cite{DTW}).

The first step of our method consists in finding the best Gaussian mixture model (GMM) fit on the demonstrations dataset and calculates the Gaussian Mixture Regression (GMR)---i.e. the generalized trajectory.
\cref{sec:method:gmmgmr} shows how to use the Jensen-Shannon (JS) divergence (\cite{lin_divergence_1991}) to fit the GMM and GMR without user input.
From the GMR, the motion is represented as a set of damped spring models; \cref{sec:method:bayes} shows how to estimate the optimal parameters of the models through Bayes optimization.
Finally, the optimal motion is computed by the attractor landscape, given initial and goal cobot joint angles. 

\subsection{Gaussian Mixture Model and Gaussian Mixture Regression}
\label{sec:method:gmmgmr}

Given a set of demonstrations, a GMM is fitted on each degree of freedom---each of the cobot's joints.
Maximum likelihood estimation of the mixture parameters is done using Expectation Maximization~(EM) (\cite{dempster_maximum_1977}).

The number of mixture model components $k$ is critical to obtaining a GMM leading to a smooth GMR.
While \textcite{Calinon2007OnLR} used the BIC criterion to determine the optimal value $k$---denoted $k^*$ in our work---\textcite{deep-MPPervez} showed that BIC overfits the dataset without a manually tuned regularization factor.
We propose a novel strategy to find $k^*$ without any manual thresholds, based on cross validation, the JS divergence, and statistical analysis.

\begin{algorithm}[t]
    \vspace{2pt}
    \KwData{demonstrations set $D$}
    \KwResult{$k^*$}
    \label{algo:method:selectionK:jessenstart}
    $s \leftarrow$ empty map\;
    \For{$k=2$ until $k=c$}{
        $res \leftarrow$ empty list\;
        \For{1 to 50}{
            Sample datapoints of $D$ in two equal sets $D_1$ and $D_2$\;
            $G_1 \leftarrow$ GMM with $k$ components fitted on $D_1$\;
            $G_2 \leftarrow$ GMM with $k$ components fitted on $D_2$\;
            Add $JSdivergence(G_1, G_2)$ to $res$\;
        }
        $m_k, s_k \leftarrow$ mean(res), std(res)\;
        $s(k) \leftarrow (m_k, s_k)$\;
    }
    \label{algo:method:selectionK:jessenend}
    $k^* \leftarrow$ component in $s$ with the lowest mean\;
    \label{algo:method:selectionK:componentstart}
    \For{{\normalfont key} $k$, {\normalfont value} $(m_k, s_k) \in s$}{
        \If{H\ref{hyp:null} is not rejected}{
            \If{H\ref{hyp:null2} is rejected {\normalfont or} $s_k < s_{k^*}$}{
                $k^* \leftarrow k$\;
            }
        }
    }
    \Return{$k^*$}
    \label{algo:method:selectionK:componentend}
    \caption{Algorithm used to determine the best number of components for the GMM.
    }
    \label{algo:method:selectionK}
\end{algorithm}

For $k=2$ until $k=c$---with $c$ the maximum number of components in the GMM---50 cross validations are performed over the demonstration dataset using the JS divergence as a measure of similarity between the GMMs generated from the train and test splits.
The mean $m_k$ and standard deviation $s_k$ of the JS divergences for each $k$ are stored in the set $s(k) \rightarrow (m_k, s_k)$.
$k^*$ is initialized as the value in $s$ with the minimum $m_k$.
For each key $k \in s$, a serie of one-tailed Welch's t-tests (\cite{Bl1947THEGO}) with $\alpha = 0.05$---i.e. there is a 5\% chance that the results occurred at random---is used to evaluate whether $k$ is a more optimal number of components than the current value of $k^*$.
First, we test if the JS divergence of $k$ is strictly greater than that of $k^*$ .
The null hypothesis H\ref{hyp:null} and alternative hypothesis H\ref{hyp:alt} are:
\begin{hyp}[H\ref{hyp:null}] \label{hyp:null}
    $k - k^* \leq 0$
\end{hyp}
\begin{hyp}[H\ref{hyp:alt}] \label{hyp:alt}
    $k^* - k < 0$
\end{hyp}
If the null hypothesis is rejected, $k$ is strictly greater than $k^*$ and $k$ is \emph{not} the optimal number of components.
If we fail to reject the null hypothesis, we then test if $k$ is strictly less than $k^*$.
The null hypothesis H\ref{hyp:null2} and alternative hypothesis H\ref{hyp:alt2} are:
\begin{hyp}[H\ref{hyp:null2}] \label{hyp:null2}
    $k^* - k \leq 0$
\end{hyp}
\begin{hyp}[H\ref{hyp:alt2}] \label{hyp:alt2}
    $k - k^* < 0$
\end{hyp}
If the null hypothesis is rejected, $k^*$ is strictly greater than $k$, and $k$ is the optimal number of components.
If we failed to reject both H\ref{hyp:null} and H\ref{hyp:null2}, no conclusions as to whether $k$ or $k^*$ is the best estimate can be drawn, and $k^*$ is set to the most stable number of components: $k^* = k$ \emph{if and only if} $s_k$ is lower than $s_{k^*}$.

The process to determine the optimal number of Gaussians is detailed in \cref{algo:method:selectionK}.

\subsection{Damped spring model}
\label{sec:method:dampedspringmodel}

\projectName{} uses the damped spring model formulated by \textcite{Ijspeert2013Nonlinear}:
%
%
\begin{equation}
    \begin{split}
        &\tau\dot{z} = \alpha_{z}(\beta_{z}(g - y) - z) + f \\
        &\dot{y} = z
    \end{split}
    \label{eq:method:dampedspring}
\end{equation}
where $\tau$ is a time constant, $f$ is the nonlinear forcing term, $\alpha_z$ and $\beta_z$ are positive constants, and $g$ is the target joint angles.
The forcing term $f$ of \cref{eq:method:dampedspring} is used to produce a specific trajectory---i.e. the GMR.
Since $f$ is a nonlinear function, it can be represented as a normalized linear combination of basis functions (\cite{Bishop}):
\begin{equation}
    f(x) = \frac{\sum_{i=1}^{N}\Psi_{i}(x) \omega_i}{\sum_{i = 1}^{N}\Psi_{i}(x)}(g-y_0)v
    \label{forcing_term}
\end{equation}
where $\Psi_i$ are fixed radial basis functions, $\omega_i$ are the weights learned during the fit, $g$ is the goal joint angles, and $v$ is the system velocity.
$N$ is the number of fixed radial basis function kernels $\Psi_{i}(x)$.
Detailed derivations, and methods to compute $\omega_i$ and the joint dynamics, are found in \textcite{Ijspeert2013Nonlinear}.

\subsection{Parameters optimization}
\label{sec:method:bayes}

For $y$ to monotonically converge towards the target $g$, the system must be critically damped on the GMR by choosing the appropriate values of $\alpha_z$ and $\beta_z$.
As shown by \textcite{Ijspeert2013Nonlinear}, $\beta_z$ can be expressed with respect to $\alpha_z$ as $4\beta_{z} = \alpha_{z}$.
Thus, only two parameters control the tracking of the reference and the stability: the number of radial basis functions $N$ and the constant $\alpha_z$.
In the previous state-of-the-art (e.g. \textcite{Pervez2018LearningTD, Ijspeert2013Nonlinear}), $\alpha_z$ and $N$ are empirically chosen by the user.
Instead, \projectName{} uses Bayesian optimization (BO) (\cite{garnett_bayesoptbook_2022}) to determine $\alpha_z$ and $N$ and avoid manual tuning.

The error to minimize is the sum of both the root mean squared error with respect to the GMR and the distance of the trajectory endpoint with respect to the goal reference:
\begin{equation}
    \begin{split}
        f(\alpha_z, N) = & \sqrt{\frac{\sum_{t=1}^{T}  \left( y(\alpha_z, N, t) - y_{G}(t) \right)^{2} }{T}} \\
        & + || y(\alpha_z, N, T) - y_{G}(T)||
    \end{split}
\end{equation}
where $y(\alpha_z, N, t)$ is the joint angles at time $t$ obtained with DMP parameters $\alpha_z$ and $N$, $y_{G}(t)$ is the GMR joint angles at time $t$, and $ || \cdot  || $ is the $l_2$-norm.
The acquisition function is the expected improvement (EI):
\begin{equation}
    EI_i (x) := E_i[f(x) - f_{i}^\ast]
\end{equation}
where $E_i[\cdot | x_{1:i}]$ indicates the expectation taken under the posterior distribution given evaluations of $f(x)$ at $x=x_1,..., x_i$.
The acquisition function retrieves the point in the search space that corresponds to the largest expected improvement and uses it for the next evaluation of the objective function $f(x)$.
The point $x_i$ minimizing the value of $f(x)$ corresponds to the optimal combination of $\alpha_z$ and $N$.
The optimization is stopped when two successive query points are equal.
\section{\uppercase{Evaluation}}

We evaluate our method in two real-world scenarios, using a 6-axis ABB GoFa CRB 15000 cobot\footnote{\url{https://new.abb.com/products/robotics/collaborative-robots/crb-15000}}---pose repeatability at the maximum reach and load is 0.05 mm.
In the first scenario, the cobot works alone to do maintenance operations on an elevator door.
This scenario is used to evaluate the stability of the method---both the parameter selection and its robustness to start and goal angle changes.
The second scenario pertains to the ease of use of our method for non-expert users: three \compagny{} workers teach the cobot a set of tasks needed to drill elevator pieces on \compagny{}'s factory line.

The desired workflow for field technicians is one where, for a given task, the cobot first learns the motion and then reproduces the motion on the factory line without having to be trained again.
Hence, for each task, a set of demonstrations is recorded by a user and the cobot learns the motion using \projectName{}.
Then, using the previously trained model, the cobot reproduces the task multiple times with different start joint angles.

To measure the cobot's accuracy in reaching the target joint angles, we measure the mean absolute error $e_j$ between the goal joint $g$ and actual end joint $t$ angles:
\begin{equation}
    e_j = \frac{\sum_{i=1}^n |t_i - g_i|}{n}
\end{equation}
with $n$ the number of joints.
To measure the quality of the reproduced motion, we use the Generalized Multiple Correlation Coefficient (GMCC) proposed by ~\textcite{GMCC}, a measure of similarities between trajectories that is invariant to linear transformations.
%
Code, datasets, and metrics can be found online.
\footnote{\url{https://github.com/SchindlerReGIS/team}}

\subsection{Door maintenance dataset}
\label{sec:evaluation:maintenance}

\begin{figure}[t]
    \centering
    \input{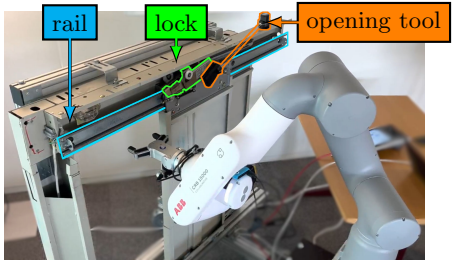}
    \caption{The cobot faces the test elevator door used for evaluation of \projectName{} in a maintenance scenario.}
    \label{fig:evaluation:maintenance}
\end{figure}

The 5 tasks of elevator door maintenance dataset: 

\begin{itemize}
    \item $T1$: open and lock the door using a custom opening tool.
          The lock is now in the middle of the rail.
    \item $T2$: grab the cleaning tool.
    \item $T3$: clean the rail while avoiding the lock.
          The cobot must aim for both ends of the rail with the brush since most dust accumulates there.
    \item $T4$: drop the cleaning tool on its support.
    \item $T5$: grab the opening tool, close the door, and combine the two pieces of the opening tool.
\end{itemize}

The maintenance setup can be seen in \cref{fig:evaluation:maintenance}.

\subsubsection{Evaluation of the parameters' repeatability}
\label{sec:evaluation:stability}

A repeatability analysis of the parameters $K$, $\alpha_z$, and $N$, is done on the data collected for the door maintenance scenario.

\emph{Repeatability of $K$ using the JS divergence}:
the method described in \cref{sec:method:gmmgmr} is run 50 times for each task in the maintenance dataset---with a minimum of 2 Gaussians and a maximum of 9.
As seen in \cref{js_stab}, the median number of Gaussians for each dataset varies only by a small standard deviation, showing that the selection of the number of Gaussian components is stable.

\begin{table}[t]
    \centering
    \caption[Jensen-Shannon divergence repeatability]{
        JS divergence repeatability over 50 runs.
    }
    \scalebox{0.75}{\begin{tabular}{l  c  c  c  c  c}
            \toprule
            Task                        & T1         & T2      & T3      & T4      & T5         \\ [0.5ex]
            \midrule
            \makecell[l]{Median nb GMM} & $5\pm0.40$ & $3\pm0$ & $4\pm0$ & $3\pm0$ & $4\pm0.27$ \\
            \bottomrule
        \end{tabular}}
    \label{js_stab}
\end{table}






\begin{table}[t]
    \centering
    \vspace{0.1cm}
    \caption{
        Comparison between grid search (GS) and Bayesian optimization (BO).
        Statistics over 50 runs.
    }
    \scalebox{0.7}{\begin{tabular}{l  c  c  c  c  c}
            \toprule
            Task                      & T1                   & T2      & T3      & T4      & T5      \\ [0.5ex]
            \midrule
            \makecell[l]{GS minimum}  & 12.84                & 73.91   & 30.88   & 36.58   & 26.44   \\
            \addlinespace[0.5em]
            \makecell[l]{BO minimum}  & \makecell[c]{$12.84$                                         \\$\pm0$}         & \makecell[c]{$74.22$\\$\pm0.61$}  & \makecell[c]{$30.98$\\$\pm0.40$} & \makecell[c]{$36.68$\\$\pm0.27$}  & \makecell[c]{$26.44$\\$\pm0.05$} \\
            \midrule
            \makecell[l]{GS time [s]} & 2348.97              & 2600.96 & 2427.34 & 2149.50 & 1848.20 \\
            \addlinespace[0.5em]
            \makecell[l]{BO median                                                                   \\time [s]} & \makecell[c]{$19.39$                                             \\$\pm3.63$} & \makecell[c]{$27.40$\\$\pm10.17$} & \makecell[c]{$12.40$\\$\pm2.40$} & \makecell[c]{$26.11$\\$\pm18.54$} & \makecell[c]{$12.53$\\$\pm1.99$} \\
            \midrule
            \makecell[l]{GS calls}    & 3750                 & 3750    & 3750    & 3750    & 3750    \\
            \addlinespace[0.5em]
            \makecell[l]{BO calls}    & \makecell[c]{$23.14$                                         \\$\pm3.88$}      & \makecell[c]{$34.20$\\$\pm10.85$} & \makecell[c]{$16.96$\\$\pm2.73$} & \makecell[c]{$32.98$\\$\pm18.97$} & \makecell[c]{$20.14$\\$\pm2.58$} \\
            \bottomrule
        \end{tabular}}
    \label{gs-table}
\end{table}

\begin{table*}[!ht]
    \centering
    \vspace{0.1cm}
    \caption{
        This table presents the training time and error metrics for the maintenance tasks--- 30 runs per task.
    }
    \scalebox{0.7}{\begin{tabular}{l l  c  c  c  c  c}
            \toprule
            Noise    & Task                               & T1              & T2              & T3              & T4              & T5              \\
            \cmidrule[0.75pt]{2-7}
                     & Number of demonstrations           & 6               & 4               & 4               & 4               & 3               \\
                     & Average demonstration duration [s] & $33.19\pm2.38$  & $36.37\pm2.28$  & $32.93\pm3.00$  & $28.17\pm3.33$  & $26.19\pm1.05$  \\
                     & Training time [s]                  & $184.08\pm2.45$ & $203.33\pm4.32$ & $185.82\pm3.07$ & $162.89\pm4.56$ & $148.59\pm4.51$ \\
            \cmidrule[0.75pt]{2-7}
                     & GMCC                               & $1.00\pm0.00$   & $1.00\pm0.00$   & $1.00\pm0.00$   & $1.00\pm0.00$   & $1.00\pm0.00$   \\
            $1$ deg  & $e_{j}$ [deg]                      & $0.14\pm0.00$   & $0.19\pm0.00$   & $0.30\pm0.00$   & $0.01\pm0.00$   & $0.05\pm0.00$   \\
            \cmidrule[0.75pt]{2-7}
                     & GMCC                               & $1.00\pm0.00$   & $1.00\pm0.00$   & $1.00\pm0.00$   & $1.00\pm0.00$   & $1.00\pm0.00$   \\
            $5$ deg  & $e_{j}$ [deg]                      & $0.14\pm0.00$   & $0.19\pm0.00$   & $0.30\pm0.00$   & $0.01\pm0.00$   & $0.05\pm0.00$   \\
            \cmidrule[0.75pt]{2-7}
                     & GMCC                               & $1.00\pm0.00$   & $1.00\pm0.00$   & $1.00\pm0.00$   & $1.00\pm0.00$   & $1.00\pm0.00$   \\
            $10$ deg & $e_{j}$ [deg]                      & $0.14\pm0.00$   & $0.19\pm0.00$   & $0.28\pm0.00$   & $0.01\pm0.00$   & $0.05\pm0.00$   \\
            \cmidrule[0.75pt]{2-7}
                     & GMCC                               & $1.00\pm0.00$   & $1.00\pm0.00$   & $1.00\pm0.00$   & $1.00\pm0.00$   & $1.00\pm0.00$   \\
            $20$ deg & $e_{j}$ [deg]                      & $0.14\pm0.00$   & $0.19\pm0.00$   & $0.28\pm0.00$   & $0.015\pm0.00$  & $0.05\pm0.00$   \\
            \bottomrule
        \end{tabular}}
    \label{data-sets}
\end{table*}

\emph{Damped spring model parameters}:
we compare BO with grid search (GS) for 50 runs per task in the maintenance dataset.
One can see in \cref{gs-table} that BO converges 100 times faster than GS and to the same global optimum.
Optimization took an average of 19.57s on an Intel Core i5 10$^{\text{th}}$ Gen and there is a reduction by a factor of at least 100 in the number of iterations needed---it should be noted that larger standard deviations in the running time are usually due to larger outliers with a median time around 20s.

In conclusion, we find that the JS divergence and BO lead to a stable selection of $K$, $\alpha_z$, and $N$, and can be used as sensible replacements for the manual tuning previously done by expert users.

\begin{figure*}[t]
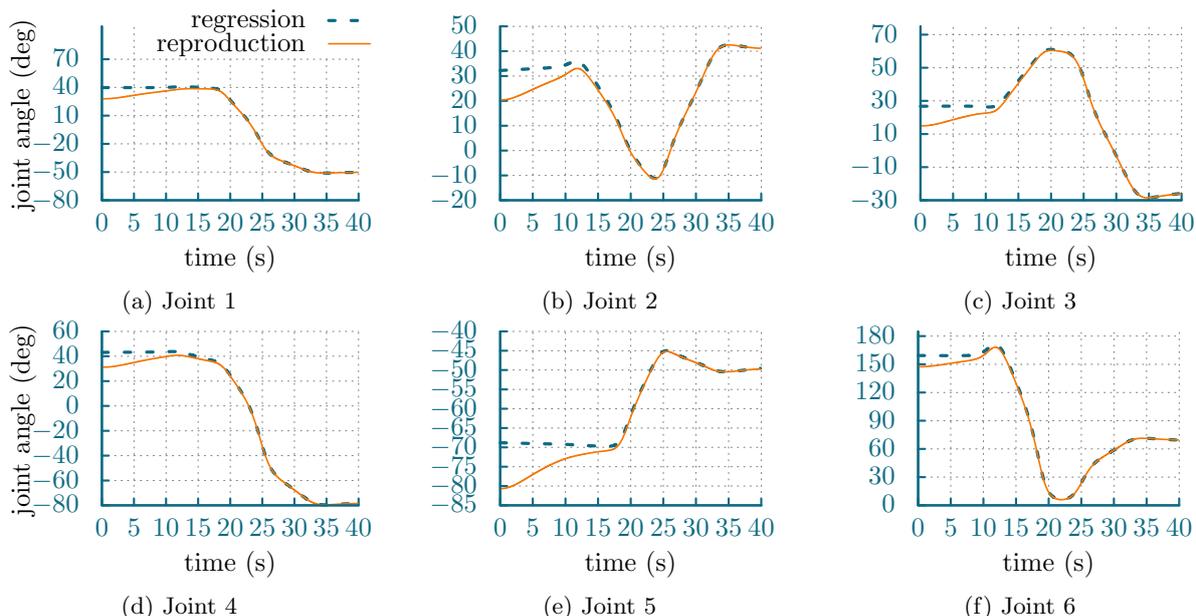

    \begin{subfigure}[b]{0.3\textwidth}
        \centering
        \begin{gnuplot}[terminal = tikz, terminaloptions = {size 5.25cm,3.6cm}]
            set tics scale 0.5
            set style line 2 lc rgb '#0d6a82' lt 1 lw 2 pt 3 ps 1
            set style line 3 lc rgb '#f57600' lt 1 lw 1.5 pt 3 ps 1
            set style line 4 lc rgb '#0d6a82' lt 2 lw 3 dashtype "."
            set border 3 ls 2
            set tics nomirror
            # #Put a grid
            set style line 12 lc rgb '#808080' lt 0 lw 1
            set grid back ls 12
            set key samplen 2
            set key at graph 1.1, 1.1
            set ytics  -80,30,71
            set yrange [-80:105]
            set ylabel "joint angle (deg)"
            set xlabel "time (s)"
            set style data points
            plot 'images/data/trajectory.dat' using 2:3 title 'regression' ls 4 with lines,\
            'images/data/trajectory.dat' using 2:9 title 'reproduction' ls 3 with lines

        \end{gnuplot}
        \vspace{-0.5cm}
        \caption{Joint 1}
    \end{subfigure}
    \hfill
    \begin{subfigure}[b]{0.3\textwidth}
        \begin{gnuplot}[terminal = tikz, terminaloptions = {size 5cm,3.6cm}]
            set tics scale 0.5
            set style line 2 lc rgb '#0d6a82' lt 1 lw 2 pt 3 ps 1
            set style line 3 lc rgb '#f57600' lt 1 lw 1.5 pt 3 ps 1
            set style line 4 lc rgb '#0d6a82' lt 2 lw 3 dashtype "."
            set border 3 ls 2
            set tics nomirror
            # #Put a grid
            set style line 12 lc rgb '#808080' lt 0 lw 1
            set grid back ls 12
            set xlabel "time (s)"
            set style data points
            plot 'images/data/trajectory.dat' using 2:4 notitle ls 4 with lines,\
            'images/data/trajectory.dat' using 2:10 notitle ls 3 with lines

        \end{gnuplot}
        \vspace{-0.5cm}
        \caption{Joint 2}
    \end{subfigure}
    \hfill
    \begin{subfigure}[b]{0.3\textwidth}
        \begin{gnuplot}[terminal = tikz, terminaloptions = {size 5cm,3.6cm}]
            set tics scale 0.5
            set style line 2 lc rgb '#0d6a82' lt 1 lw 2 pt 3 ps 1
            set style line 3 lc rgb '#f57600' lt 1 lw 1.5 pt 3 ps 1
            set style line 4 lc rgb '#0d6a82' lt 2 lw 3 dashtype "."
            set border 3 ls 2
            set tics nomirror
            # #Put a grid
            set style line 12 lc rgb '#808080' lt 0 lw 1
            set grid back ls 12
            set ytics  -30,20,70
            set yrange[-30:75]
            set xlabel "time (s)"
            set style data points
            plot 'images/data/trajectory.dat' using 2:5 notitle ls 4 with lines,\
            'images/data/trajectory.dat' using 2:11 notitle ls 3 with lines

        \end{gnuplot}
        \vspace{-0.5cm}
        \caption{Joint 3}
    \end{subfigure}

    \begin{subfigure}[b]{0.3\textwidth}
        \begin{gnuplot}[terminal = tikz, terminaloptions = {size 5.25cm,3.6cm}]
            set tics scale 0.5
            set style line 2 lc rgb '#0d6a82' lt 1 lw 2 pt 3 ps 1
            set style line 3 lc rgb '#f57600' lt 1 lw 1.5 pt 3 ps 1
            set style line 4 lc rgb '#0d6a82' lt 2 lw 3 dashtype "."
            set border 3 ls 2
            set tics nomirror
            # #Put a grid
            set style line 12 lc rgb '#808080' lt 0 lw 1
            set grid back ls 12

            set ylabel "joint angle (deg)"
            set xlabel "time (s)"
            set style data points
            plot 'images/data/trajectory.dat' using 2:6 notitle ls 4 with lines,\
            'images/data/trajectory.dat' using 2:12 notitle ls 3 with lines

        \end{gnuplot}
        \vspace{-0.5cm}
        \caption{Joint 4}
    \end{subfigure}
    \hfill
    \begin{subfigure}[b]{0.3\textwidth}
        \begin{gnuplot}[terminal = tikz, terminaloptions = {size 5cm,3.6cm}]
            set tics scale 0.5
            set style line 2 lc rgb '#0d6a82' lt 1 lw 2 pt 3 ps 1
            set style line 3 lc rgb '#f57600' lt 1 lw 1.5 pt 3 ps 1
            set style line 4 lc rgb '#0d6a82' lt 2 lw 3 dashtype "."
            set border 3 ls 2
            set tics nomirror
            # #Put a grid
            set style line 12 lc rgb '#808080' lt 0 lw 1
            set grid back ls 12

            set xlabel "time (s)"
            set style data points
            plot 'images/data/trajectory.dat' using 2:7 notitle ls 4 with lines,\
            'images/data/trajectory.dat' using 2:13 notitle ls 3 with lines

        \end{gnuplot}
        \vspace{-0.5cm}
        \caption{Joint 5}
    \end{subfigure}
    \hfill
    \begin{subfigure}[b]{0.3\textwidth}
        \begin{gnuplot}[terminal = tikz, terminaloptions = {size 5cm,3.6cm}]
            set tics scale 0.5
            set style line 2 lc rgb '#0d6a82' lt 1 lw 2 pt 3 ps 1
            set style line 3 lc rgb '#f57600' lt 1 lw 1.5 pt 3 ps 1
            set style line 4 lc rgb '#0d6a82' lt 2 lw 3 dashtype "."
            set border 3 ls 2
            set tics nomirror
            # #Put a grid
            set style line 12 lc rgb '#808080' lt 0 lw 1
            set grid back ls 12
            set ytics  0,30,180
            set yrange[0:185]
            set xlabel "time (s)"
            set style data points
            plot 'images/data/trajectory.dat' using 2:8 notitle ls 4 with lines,\
            'images/data/trajectory.dat' using 2:14 notitle ls 3 with lines

        \end{gnuplot}
        \vspace{-0.5cm}
        \caption{Joint 6}
    \end{subfigure}

    \caption{Reproduction and regression joint angles evolution on an example of the $T1$ task.
        Start joint angles of the reproduction are computed by adding Gaussian noise with 20deg standard deviation on the regression initial joint angles.
        The reproduced trajectory reproduces the regression's motion and reach the target joint angles.
    }
    \label{fig:evaluation:noise-on-references}
\end{figure*}

\subsubsection{Adaptability}

\cref{data-sets} shows the number of demonstrations recorded for each task, with the average training times and error metrics.

The complexity of DTW is $\mathcal{O}((M-1)L^{2})$, with $M$ the number of demonstrations in the dataset and $L$ the longest demonstration length,
GMM and GMR are $\mathcal{O}(KBD^{3})$ where $B$ is the number of datapoints in the dataset, $D$ the data dimensionality, and $K$ the number of GMM components.
The Gaussian Process of the BO is $\mathcal{O}(R^{3})$ with $R$ being the number of function evaluations---\cref{gs-table} shows that $R$ is at worse around $34.20\pm10.85$.
As seen in \cref{data-sets}, in the maintenance scenario, the maximum training time is under 4min---the longest training time is $203.33\pm4.32$s for $T2$.

Once a model is trained, the accuracy of the reproduced trajectory is evaluated by computing 30 reproductions and calculating the GMCC between the reproduced trajectory and the GMR.
For each reproduction, the target joint angles are the same as the last joint angles of the GMR, and the start joint angles are the same as the GMR, with the addition of a zero mean Gaussian noise with standard deviation of $1$, $5$, $10$, and $20$ degrees.
For each noise value, the average GMCC and $e_j$ per task are shown in \cref{data-sets}.
One can see that, regardless of the noise value, GMCCs and $e_j$ are very close to $1$ and $0$ respectively, showing that the cobot accurately reproduces the demonstrated motion and reaches the target joint angles.
E.g., \cref{fig:evaluation:noise-on-references} shows the regression trajectory and reproduction per joint with Gaussian noise with standard deviation of 20 deg: the trajectory of each joint is conserved regardless of the noise added to the start joint angles.

\subsection{Field tests and user study}

\begin{table*}[!t]
    \centering
    \vspace{5pt}
    \caption{
        This table presents error metrics for the factory scenario---dataset consisted of 3 to 4 demonstrations.
    }
    \scalebox{0.75}{\begin{tabular}{l  c  c  c  c}
            \toprule
            Task                       & F1            & F2            & F3            & F4            \\ [0.5ex]
            \midrule
            Number of reproductions     & 31            & 30            & 33            & 20            \\
            GMCC                       & $0.99\pm0.02$ & $0.99\pm0.00$ & $0.99\pm0.01$ & $1.00\pm0.00$ \\
            Joints error $e_{j}$ [deg] & $0.00\pm0.00$ & $0.02\pm0.02$ & $0.04\pm0.05$ & $0.02\pm0.02$ \\
            \bottomrule
        \end{tabular}}

    \label{factory-results}
\end{table*}

To validate that the cobot can be used by non-expert users in a professional setting, we conducted field tests at the \compagny{} headquarters with three \compagny{} field workers working on the production line.
None of the workers had worked with a cobot before.
To ensure realism of the tasks, the field workers designed four test scenarios that would reduce their workload \emph{if} the robot can easily be taught how to perform the task:
\begin{itemize}
    \item $F1$: find a metal piece, grab and place it on a drilling machine.
    \item $F2$: find a metal piece, grab and place it on a drilling machine while avoiding an obstacle.
    \item $F3$: find a metal frame, grab and place it on a drilling machine while rotating the piece.
    \item $F4$: find a wooden plank, grab one side while the worker grabs the other, and place it together on a drilling machine.
\end{itemize}
A custom app on a smartphone was used by the workers to interact with the robot in an intuitive manner.
The app consists of two main pages: one to record demonstrations and train a model, and another page to give the robot a target position and start the task reproduction.
After a short training on how to use the app, three to five demonstrations were recorded per user, per task.
To calculate the metric, each task was reproduced around 10 times per user, apart from $F4$ where only two users participated---hence $20$ reproductions.
Detection of the different objects is done using template matching (\cite{Brunelli2009TemplateMT}).

\cref{factory-results} shows the GMCC and $e_j$ for all tasks, calculated for 30 reproductions of the motion for each task.
The error metrics results are similar to the ones presented in \cref{sec:evaluation:maintenance}, with GMCC averaging $0.99$ and $e_j$ of $0$ deg; demonstrating accurate task reproduction in a realistic scenario.

\begin{figure*}[t]
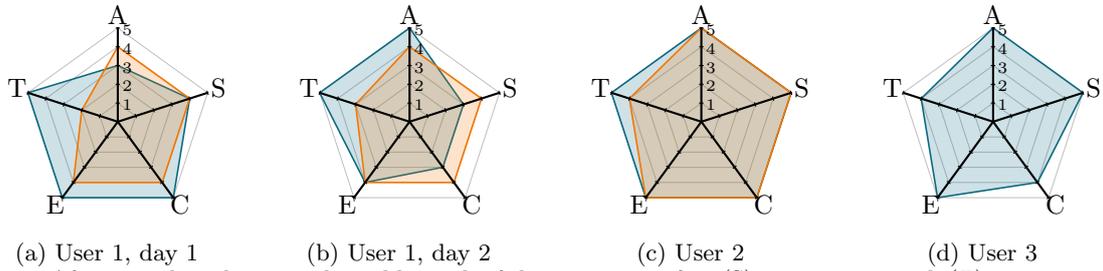

    \centering
    \begin{subfigure}[b]{0.2\textwidth}
        \begin{gnuplot}[terminal = tikz, terminaloptions = {size 3.5cm,5cm}]
            set style line 2 lc rgb '#0d6a82' lt 1 lw 1.5
            set style line 3 lc rgb '#f57600' lt 1 lw 1.5
            set style line 4 lc rgb '#0d6a82' lt 2 lw 1.5
            set style line 5 lc rgb '#84a4b3' lt 2 lw 1.5
            set datafile separator comma
            set spiderplot
            set style spiderplot fs transparent solid 0.2 border lw 1 pt 6 ps 2
            set for [p=1:5] paxis p range [0:5]
            set for [p=2:5] paxis p tics format ""
            set paxis 1 tics font ",6"
            set grid spider lt black lc "grey" lw 0.5 back

            plot for [i=2:6] "images/data/user1.dat" using i title columnhead ls 2,\
            newspiderplot, \
            for [i=2:6] "images/data/user1_colab.dat" using i title columnhead  ls 3

        \end{gnuplot}
        \vspace{-1.5cm}
        \caption{User 1, day 1}
    \end{subfigure}
    \hfil
    \begin{subfigure}[b]{0.2\textwidth}
        \begin{gnuplot}[terminal = tikz, terminaloptions = {size 3.5cm,5cm}]
            set style line 2 lc rgb '#0d6a82' lt 1 lw 1.5
            set style line 3 lc rgb '#f57600' lt 1 lw 1.5
            set style line 4 lc rgb '#0d6a82' lt 2 lw 1.5
            set style line 5 lc rgb '#84a4b3' lt 2 lw 1.5

            set datafile separator comma
            set spiderplot
            set style spiderplot fs transparent solid 0.2 border
            set for [p=1:5] paxis p range [0:5]
            set for [p=2:5] paxis p tics format ""
            set paxis 1 tics font ",6"
            set grid spider lt black lc "grey" lw 0.5 back
            plot for [i=2:6] "images/data/user4.dat" using i title columnhead ls 2,\
            newspiderplot, \
            for [i=2:6] "images/data/user4_colab.dat" using i title columnhead  ls 3

        \end{gnuplot}
        \vspace{-1.5cm}
        \caption{User 1, day 2}
    \end{subfigure}
    \hfil
    \begin{subfigure}[b]{0.2\textwidth}
        \begin{gnuplot}[terminal = tikz, terminaloptions = {size 3.5cm,5cm}]
            set style line 2 lc rgb '#0d6a82' lt 1 lw 1.5
            set style line 3 lc rgb '#f57600' lt 1 lw 1.5
            set style line 4 lc rgb '#0d6a82' lt 2 lw 1.5
            set style line 5 lc rgb '#84a4b3' lt 2 lw 1.5

            set datafile separator comma
            set spiderplot
            set style spiderplot fs transparent solid 0.2 border
            set for [p=1:5] paxis p range [0:5]
            set for [p=2:5] paxis p tics format ""
            set paxis 1 tics font ",6"
            set grid spider lt black lc "grey" lw 0.5 back
            plot for [i=2:6] "images/data/user2.dat" using i title columnhead ls 2,\
            newspiderplot, \
            for [i=2:6] "images/data/user2_colab.dat" using i title columnhead  ls 3

        \end{gnuplot}
        \vspace{-1.5cm}
        \caption{User 2}
    \end{subfigure}
    \hfil
    \begin{subfigure}[b]{0.2\textwidth}
        \begin{gnuplot}[terminal = tikz, terminaloptions = {size 3.5cm,5cm}]
            set style line 2 lc rgb '#0d6a82' lt 1 lw 1.5
            set style line 3 lc rgb '#f57600' lt 1 lw 1.5
            set style line 4 lc rgb '#0d6a82' lt 2 lw 1.5
            set style line 5 lc rgb '#84a4b3' lt 2 lw 1.5

            set datafile separator comma
            set spiderplot
            set style spiderplot fs transparent solid 0.2 border
            set for [p=1:5] paxis p range [0:5]
            set for [p=2:5] paxis p tics format ""
            set paxis 1 tics font ",6"
            set grid spider lt black lc "grey" lw 0.5 back
            plot for [i=2:6] "images/data/user3.dat" using i title columnhead ls 2

        \end{gnuplot}
        \vspace{-1.5cm}
        \caption{User 3}
    \end{subfigure}

    \caption{After use, the cobot is evaluated by each of the users on: safety (S), easiness to teach (T), entertainment (E), reaching the target joint angles (A), and task completion (C).
        In blue the results for the task F1, F2, and F3, and in red the results for the collaborative task F4.
        One can see the collaborative task, where the user carries a piece of wood with the robot, is more difficult than other task were the cobot works next to the user.
    }
    \label{fig:survey}
\end{figure*}


The field tests were conducted over two days and, at the end of each day, the workers answered a questionnaire to evaluate the cobot's performance.
In the survey, users rate the following statements on a scale from 1 to 5, corresponding to strongly disagree, disagree, neutral, agree, and strongly agree:
\begin{itemize}
    \item The cobot learned the correct motion.
    \item I felt safe operating the cobot.
    \item The cobot reached the goal point accurately.
    \item Teaching the cobot a motion was simple.
    \item Teaching the cobot a motion was entertaining.
\end{itemize}
The radar plots in \cref{fig:survey} present survey results.
While users showed satisfaction with the cobot's precision and motion performance, the complexity of holding the beam and moving the cobot while showing the motion in F4 led to a lower score for easiness of teaching compared to other tasks.

\section{\uppercase{Limitations and future work}}

\projectName{} doesn't consider elements of the environment during the motion.
This create confusion for the workers not understanding why the cobot does not avoid obstacles, making it harder for them to trust the cobot.
Future work will look at integrating visual information through cameras to update the motion depending on the environment.

Another way that \projectName{} could be improved is by being able to update the attractor landscape of a motion incrementally.
Future work will look into making the process incremental, giving workers the ability to correct existing motions learned by the cobot.

\section{\uppercase{Summary}}

A method to learn motions from demonstrations requiring no manual parameter tuning has been developed.
Given a set of demonstrations aligned in time, the motion is generalized using GMM and the reference trajectory is extracted with GMR.
Since BIC criterion can lead to over-fitting of the GMM, it is proposed to instead use the Jensen-Shannon divergence to determine the optimal number of GMM components.
The cobot DOFs are represented as damped spring models and the forcing term is learned to adapt the motion to different start and goal joint poses.
Parameters of the spring model are found using Bayesian optimization.

\projectName{} is extensively evaluated in two field tests where the cobot performs tasks related to elevator door maintenance, and works in realistic scenarios with \compagny{} field workers.
The precision in joint angles and motion reproduction quality are evaluated, and the experiments show that the cobot accurately reproduces the motions---GMCC and mean average error for the final joint angles are around $1$ and $0$ respectively.
Furthermore, feedback collected from the field workers shows that the cobot is positively accepted since it is easy to teach and easy to use.

\printbibliography 



\end{document}